\documentclass[final,5p,times,twocolumn]{elsarticle}

\usepackage{amssymb}
\usepackage{amsmath}
\usepackage{booktabs}

\journal{Knowledge-Based Systems}

\begin{document}

\begin{frontmatter}



\title{Knowledge-Guided Local-to-Global Anomaly Learning for \\Longitudinal Medication Trajectories} 


\author{Wendi Ji} 

\affiliation{organization={Department of Intelligent Science and Information Law, East China University of Political Science and Law},
            addressline={555 Longyuan Road, Songjiang University Town}, 
            city={Shanghai},
            postcode={201620}, 
            country={China}}

\begin{abstract}
Longitudinal medication trajectories capture evolving drug combinations and
provide a basis for identifying potential care-related risks, with anomalous
deviations often confined to a few visits or treatment episodes. However,
generic anomaly supervision struggles to distinguish plausible medication
changes from suspicious semantic deviations, while global modeling can
dilute local signals and account-level supervision alone leaves anomaly
locations underdetermined. We propose a knowledge-guided local-to-global
(L2G) anomaly learning framework for joint account-level anomaly detection
and visit-level localization. Within this framework, L2G-Joint constructs
knowledge-guided proxy anomalies at individual-visit, treatment-episode,
and contiguous-segment scales, with knowledge-consistent plausible negatives
defining a supervision boundary between reasonable variations and suspicious
deviations. The model assigns a local anomaly score to each scoreable visit
and derives the account-level score through selective Top-$\alpha$
aggregation. Account- and visit-level objectives jointly constrain the same
scores, making localization integral to learning global anomaly decisions
rather than merely a post-hoc explanation. Experiments on proxy benchmarks
constructed from medication records in two cities compare L2G-Joint against
12 literature-based baseline configurations and two controlled variants. On
City-A, L2G-Joint achieves a Macro AP of 0.7844, a Macro AUROC of 0.7707,
and a Localization Macro AP of 0.7215, representing relative improvements of
2.4\%, 3.6\%, and 9.2\% over the strongest respective comparators. Ablation
and cross-city analyses further support knowledge-guided supervision and
selective local aggregation, while showing that explicit local supervision
improves localization but its additional benefit for account-level detection
varies across cohorts.
\end{abstract}


\end{frontmatter}




\section{Introduction}

Longitudinal healthcare records capture how diagnoses, treatments, medications, and other clinical events evolve over repeated encounters, providing a rich basis for identifying deviations from expected care patterns. Such deviations can be clinically or operationally important, and anomaly detection has therefore been increasingly applied to longitudinal clinical processes, healthcare utilization, and insurance records \cite{huang2012clinical,hu2012utilization,niu2022anomalous,niu2024ehrbert}. Yet longitudinal medical data differ fundamentally from conventional independent observations or fixed-dimensional time series. Patient trajectories are heterogeneous, treatment pathways evolve over time, and clinically meaningful irregularities may occur only within a small part of an otherwise plausible history \cite{perer2015care,muyama2024clinicalpathways,gebeyehu2026rare}. Medication trajectories provide a particularly representative setting: each visit may contain multiple co-occurring medications, while changes across visits reflect evolving therapeutic decisions and clinical context. Preserving this longitudinal structure is important because clinically relevant deviations may arise from a small number of altered medications, a temporally coherent treatment episode, or a contiguous segment of visits rather than from the trajectory as a whole. Detecting anomalies in such records therefore requires reasoning about both the overall trajectory and the localized evidence embedded within it.

Existing studies provide several foundations for longitudinal anomaly detection. Clinical-process and EHR methods identify deviations from recurring temporal or contextual patterns \cite{huang2015local,niu2022anomalous,niu2024ehrbert}, while deep sequential models characterize abnormality through one-class learning, reconstruction, prediction, or contextual discrepancy \cite{ruff2018deepsvdd,tuli2022tranad,carmona2022ncad}. When verified anomalies are unavailable, methods such as NCAD and AnomalyBERT further construct synthetic perturbations to provide anomaly supervision \cite{carmona2022ncad,jeong2023anomalybert}. However, these anomaly definitions are largely generic and are not designed to distinguish clinically plausible medication changes from semantically inconsistent ones. For longitudinal medication trajectories, the first challenge is therefore to construct anomaly supervision that reflects domain-relevant deviations rather than merely artificial or distributional irregularities.
A second challenge concerns the connection between local and global anomaly evidence. Prior work has considered local clinical deviations \cite{huang2015local}, multi-scale event-sequence anomalies \cite{wang2021oc4seq}, and position-level anomaly prediction \cite{jeong2023anomalybert}. Yet in medication trajectories, only a small subset of visits may contain the deviation, while the final decision is often required at the account level. Global aggregation can dilute sparse signals, whereas account-level supervision alone leaves the responsible visits underdetermined. This motivates an explicit local-to-global formulation in which visit-level anomaly evidence is learned and selectively aggregated to form the account-level decision, with local supervision further constraining which visits should support that decision.

To address these two challenges, we propose a
\textbf{knowledge-guided local-to-global anomaly learning framework}
for longitudinal medication trajectories. First, we construct
multi-granularity proxy anomalies under medication knowledge constraints,
covering sparse medication changes, temporally coherent treatment episodes,
and contiguous visit segments. Unlike generic synthetic corruptions, our
framework introduces knowledge-consistent plausible variations together
with suspicious semantic deviations, enabling the model to learn an anomaly
boundary that distinguishes meaningful medication evolution from
potentially abnormal changes.
Beyond anomaly supervision, we further formulate anomaly detection as a
local-to-global evidence learning problem. Each scoreable visit is treated
as a local anomaly evidence unit, and account-level decisions are derived
by selectively aggregating informative local evidence rather than relying
on an independent global representation. The proposed L2G-Joint model jointly
optimizes account-level detection and visit-level localization using the
same visit-level anomaly scores, allowing localization supervision to
identify which local deviations should support the global decision. Thus,
visit-level localization is not a post-hoc interpretation of anomaly
predictions, but an integral component of learning the evidence underlying
account-level detection.
The proposed framework addresses anomaly detection as a two-stage reasoning
process. Knowledge-guided proxy construction defines what constitutes a
meaningful deviation in medication trajectories, while local-to-global
learning determines how such deviations should contribute to an account-level
decision. By coupling anomaly definition with evidence attribution, the
framework provides both supervision aligned with medication semantics and
localized evidence supporting global anomaly assessment.

Extensive experiments are conducted on knowledge-guided proxy benchmarks
constructed from longitudinal medication records in two independent cities.
We compare the proposed framework with native anomaly detection methods,
proxy-adapted variants, and controlled models. Results demonstrate that
knowledge-aligned anomaly supervision substantially improves anomaly
detection performance, while local-to-global joint learning provides
consistent benefits for both account-level detection and visit-level
localization. Additional analyses further validate the contributions of
plausible negatives, selective aggregation, and cross-city robustness.

The main contributions of this work are summarized as follows:

\begin{itemize}

\item We formulate a knowledge-guided anomaly supervision framework for
longitudinal medication trajectories, where multi-granularity proxy anomalies
and plausible negatives distinguish clinically consistent medication
variations from suspicious semantic deviations.

\item We propose a local-to-global anomaly learning framework that derives
account-level decisions from selectively aggregated visit-level anomaly
evidence, while jointly learning anomaly localization through shared visit
scores.

\item We introduce a joint perspective that connects anomaly supervision
alignment with evidence-level learning, enabling models to learn not only
whether a trajectory is suspicious but also which local deviations support
the global decision.

\item We conduct extensive evaluations on two independent city cohorts,
covering account-level anomaly detection and visit-level localization, and
provide systematic analyses of supervision alignment, plausible negatives,
and local-to-global aggregation.
\end{itemize}

\section{Related Work}

This section reviews prior studies from three perspectives most relevant to our work: longitudinal clinical-event anomaly detection, deep anomaly modeling for sequential and event data, and synthetic or knowledge-guided anomaly learning.

\subsection{Anomaly Detection in Longitudinal Clinical Event Sequences}
\label{sec:related-clinical-anomaly}

Longitudinal clinical records exhibit recurring temporal and process regularities across diagnoses, treatments, examinations, and other care activities, providing a natural basis for identifying deviations from expected care patterns. \citet{bouarfa2012workflow}, for example, derived a consensus workflow from surgical activity logs and detected procedures that deviated from common sequence structures. Process-oriented studies further showed that recurrent care patterns can be recovered from routinely collected event logs: \citet{caron2014process} analyzed variations in historical clinical pathway executions, while \citet{perer2015care} extracted diagnosis and treatment pathways while accounting for concurrent events, heterogeneous event granularity, and temporal context. Recent reviews confirm that clustering, process mining, and increasingly machine-learning and deep-learning techniques have been widely used to discover longitudinal regularities in healthcare data \cite{muyama2024clinicalpathways,bhanja2026ehrsurvey,gebeyehu2026rare}. Such regularities have supported applications ranging from clinical-process monitoring to anomaly screening in healthcare utilization and insurance claims \cite{hu2012utilization,suesserman2023procedure,settipalli2023wmtdbc,demeulemeester2025explainable}. These studies establish the basic premise of longitudinal clinical anomaly detection: atypical trajectories can be identified relative to recurring patterns in historical medical-event data.

At the same time, longitudinal care exhibits substantial heterogeneity across patients and treatment pathways, making a single guideline or consensus trajectory an insufficient definition of normality \cite{muyama2024clinicalpathways,perer2015care}. This motivated a shift from fixed reference pathways toward data-driven normality learned from observed patient histories. \citet{huang2012clinical} explicitly avoided assuming a predefined clinical-process model and instead inferred common careflow patterns from historical traces, allowing unusual trajectories to be identified relative to population-derived regularities. Later EHR studies extended this principle to more general event sequences. \citet{niu2022anomalous} characterized normal sequence structure through higher-order event dependencies and event durations, whereas \citet{niu2024ehrbert} used sequential masked-token prediction to learn contextual dependencies among clinical events. These developments reflect a broader transition from rule- or pathway-based conformity toward adaptive normality learned from heterogeneous longitudinal data \cite{bhanja2026ehrsurvey}.

Clinically relevant anomalies may also arise from qualitatively different forms of deviation. Process and temporal irregularities include unexpected, early, delayed, or absent events \cite{huang2015local}, as well as violations of higher-order event dependencies or duration patterns \cite{niu2022anomalous}. Other anomalies arise from contextual incompatibility among healthcare contents: \citet{suesserman2023procedure}, for instance, identified potentially unwarranted procedures by modeling their compatibility with diagnoses, patient characteristics, and other procedures. A further class concerns atypical utilization or behavioral patterns, such as healthcare use that deviates from patient-specific expectations \cite{hu2012utilization} or unusual provider-level utilization profiles \cite{demeulemeester2025explainable}. Claims-oriented studies have also directly formulated such deviations as unsupervised fraud-detection problems \cite{settipalli2023wmtdbc}. Recent reviews similarly emphasize the rarity and heterogeneity of abnormal medical events as persistent challenges for data-driven healthcare anomaly detection \cite{gebeyehu2026rare}. Thus, longitudinal healthcare anomalies cannot be reduced to a single predefined pattern, but may reflect distinct temporal, contextual, and behavioral deviations.

Longitudinal medication records represent a specific instance of this broader problem, in which medication use evolves across successive patient visits. Recent evidence further suggests that preserving event order can reveal clinically relevant information that is lost when longitudinal records are reduced to unordered code sets \cite{edelman2026trajectories}. Rather than treating medication records as isolated prescriptions or aggregated utilization statistics, we therefore study them as longitudinal medication trajectories and investigate anomaly detection within this clinical-event setting.

\subsection{Deep Anomaly Modeling for Sequential and Event Data}

Deep anomaly modeling for sequential data involves three closely related issues: how normal patterns are characterized, how different sequence structures are represented, and how sequence-level decisions can be linked to localized anomaly evidence. Existing methods accordingly span one-class, reconstruction-, prediction-, and dependency-based normality modeling, as well as discrete event-sequence representation and weakly supervised localization \cite{pang2021deepad}. These directions provide the methodological basis for modeling anomalies in longitudinal medication trajectories.

A major distinction among deep anomaly detection methods lies in how deviations from normal sequential behavior are quantified. One-class approaches directly learn compact representations of normal data. \citet{ruff2018deepsvdd} proposed Deep SVDD, which maps normal samples toward a compact hypersphere and measures abnormality by their distance from the learned center. Extending one-class learning to temporal data, \citet{shen2020thoc} introduced THOC, which combines multi-scale recurrent representations with multiple one-class hyperspheres to characterize temporal dynamics at different resolutions. Other methods learn normality through reconstruction or probabilistic sequence modeling. \citet{su2019omnianomaly} modeled multivariate time series with a stochastic recurrent network and derived anomaly scores from reconstruction probabilities, while \citet{tuli2022tranad} used Transformer-based modeling, self-conditioning, and adversarial training to reconstruct observations in their temporal context. Beyond reconstruction, \citet{xu2022anomalytransformer} defined anomalies through association discrepancy, capturing differences between the temporal associations of normal and anomalous observations. These approaches illustrate a progression from distance-based normality modeling to reconstruction-, probability-, and dependency-based anomaly criteria. However, they are primarily designed for conventional multivariate time series, where each time step is represented by a fixed-dimensional numerical vector.

Discrete event sequences require a different form of sequence modeling because abnormality may depend on event identity, semantic content, and sequential organization rather than numerical deviation alone. \citet{du2017deeplog} modeled system logs as event sequences and detected anomalies through unexpected next-event predictions. \citet{meng2019loganomaly} further incorporated semantic representations of log templates and quantitative patterns, showing that event identifiers alone may be insufficient to characterize abnormal behavior. Moving beyond next-event prediction, \citet{wang2021oc4seq} combined one-class learning with multi-scale recurrent representations to capture anomalies at the event, subsequence, and whole-sequence levels, while \citet{zhang2022cat} introduced a content-aware Transformer that jointly represents event identity and event content. Recent work has further considered generalization across heterogeneous event streams: \citet{guo2024logformer} used Transformer pre-training and adapter tuning for cross-domain log anomaly detection, and \citet{guan2025feadseq} developed personalized federated learning for discrete event-sequence anomaly detection under heterogeneous local distributions. Overall, these studies progressively extend anomaly modeling from numerical observations to semantic and structured discrete events. Nevertheless, they generally assume that each sequence position corresponds to a single event or event representation; longitudinal records in which each temporal unit contains multiple co-occurring items therefore require an additional level of structural modeling.

\subsection{Synthetic and Knowledge-Guided Anomaly Learning}

When verified anomalies are scarce, a central question is how to obtain informative anomaly supervision without relying on large labeled abnormal datasets. Early anomaly-informed approaches showed that even limited abnormal examples can improve detection beyond purely normality-based learning. \citet{hendrycks2019oe} introduced Outlier Exposure, demonstrating that auxiliary out-of-distribution samples can improve generalization to unseen anomalies. \citet{pang2019devnet} proposed DevNet to learn anomaly scores directly from a small number of labeled anomalies, while \citet{ruff2020deepsad} extended one-class learning to explicitly incorporate labeled abnormal examples. These studies established the value of anomaly supervision, but still assume access to external outliers or verified anomalies, which are often difficult to obtain in real applications.

This limitation motivated approaches that construct artificial supervision directly from available normal data. \citet{goyal2020drocc} generated adversarial negative samples around the normal-data manifold to learn a discriminative boundary, while transformation-based methods created proxy tasks by deliberately perturbing normal samples. \citet{bergman2020goad} used random affine transformations for self-supervised anomaly representation learning, and \citet{li2021cutpaste} showed that synthetic visual defects could induce representations that transfer to previously unseen real anomalies. \citet{zavrtanik2021draem} further used simulated anomalies to train a discriminative model for both detection and localization. A key insight from this line of work is that synthetic anomalies need not reproduce real anomalies exactly; their value lies in constructing informative violations of normal structure. This shifts the methodological focus from acquiring abnormal labels to designing useful proxy anomalies.

Synthetic anomaly construction has subsequently become an explicit strategy for sequential data. \citet{carmona2022ncad} injected generic anomalies into time-series windows to learn contextual normal--abnormal boundaries, while \citet{jeong2023anomalybert} used several degradation mechanisms to generate temporal outliers and directly supervise position-level anomaly prediction. \citet{xu2024couta} further introduced generated native anomalies to compensate for insufficient anomaly knowledge in one-class learning. More recently, \citet{tang2024pcrta} explicitly distinguished perturbation types and used perturbation pseudo-labels to guide representation learning, emphasizing that different synthetic transformations provide different amounts of anomaly information. These developments move sequential anomaly detection from generic corruption toward more deliberate perturbation design, where the structure of the proxy task itself becomes part of the anomaly-learning method.

Recent work has begun to incorporate prior or learned knowledge into this process. \citet{qiu2024augpaste} used lesion priors to guide the synthesis of abnormal medical images, illustrating how domain-relevant cues can make synthetic anomalies more meaningful. For time-series data, \citet{xia2026eastkb} combined enhanced anomaly simulation with a triplet knowledge base that maintains general, normal, and abnormal representations. These studies mark a transition from generic perturbations toward knowledge-enhanced anomaly learning. Our work follows this direction but uses knowledge at an earlier stage of the pipeline: external medication knowledge and longitudinal trajectory structure directly constrain which modifications are admissible when constructing proxy anomalies. Consequently, the generator defines not only synthetic abnormal trajectories but also the visits directly affected by each perturbation, yielding both trajectory-level and visit-level proxy supervision. Knowledge-guided synthesis therefore serves as a mechanism for constructing structured, multi-granularity anomaly supervision rather than merely as a form of data augmentation.

\section{Method}
\label{sec:method}

\begin{figure*}[t]
    \centering
    \includegraphics[width=0.8\textwidth]{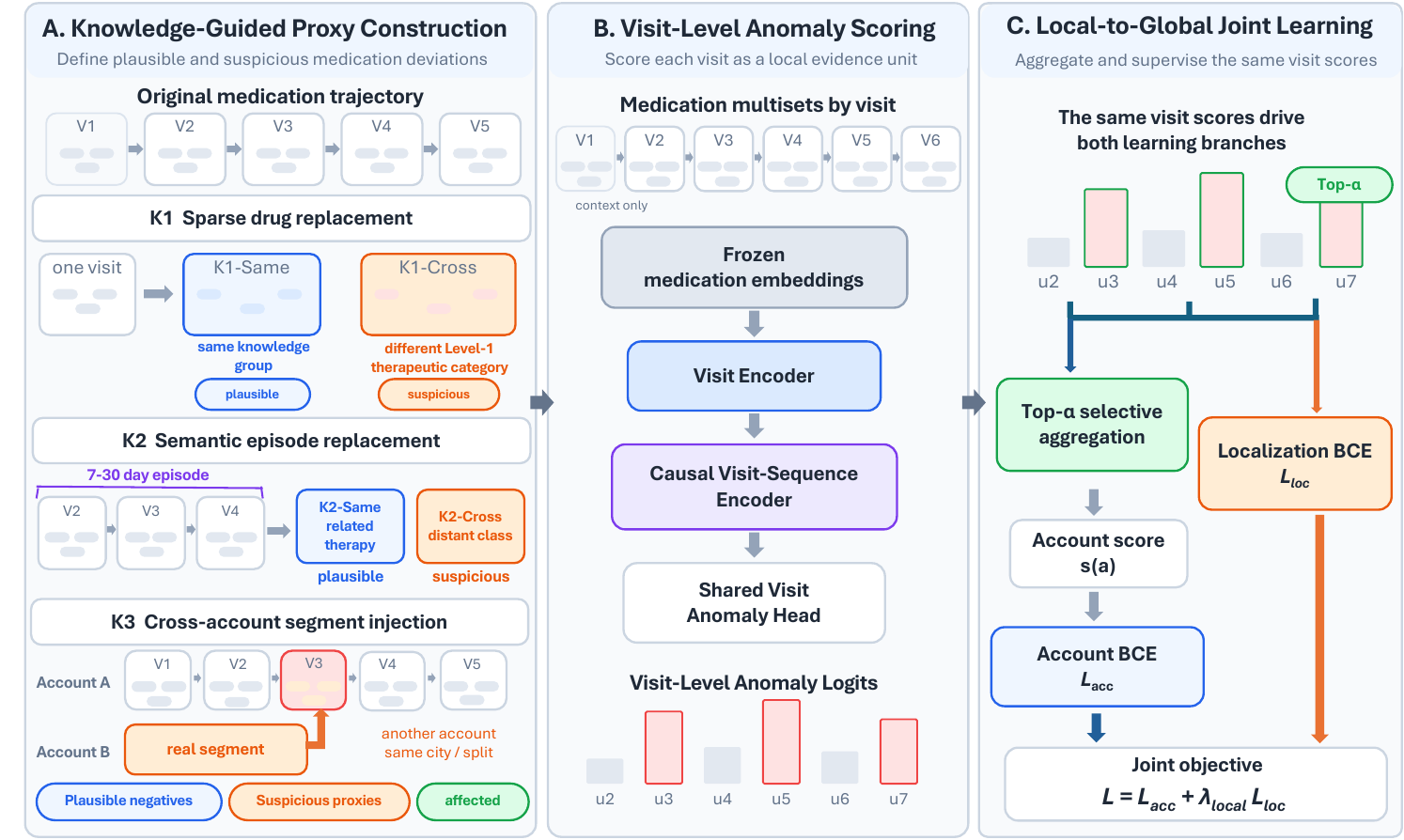}
\caption{
Overview of the proposed knowledge-guided proxy anomaly construction and
local-to-global anomaly learning framework.
Knowledge-guided operators construct plausible and suspicious medication
deviations at the medication-occurrence, treatment-episode, and
contiguous-segment levels. The model learns visit-level anomaly evidence
and selectively aggregates high-scoring visits through Top-$\alpha$
aggregation to obtain account-level anomaly logits.
Account-level and visit-level objectives jointly supervise the same
visit-level anomaly logits, coupling local anomaly localization with
global account-level detection.
}
    \label{fig:framework}
\end{figure*}

We propose a knowledge-guided local-to-global anomaly learning framework
for longitudinal medication trajectories. The framework addresses two
coupled challenges: identifying suspicious accounts and locating the
corresponding visits, in the absence of reliable anomaly annotations.
We construct multi-granularity proxy anomalies
through knowledge-guided perturbations of medication trajectories,
thereby providing account-level supervision and positive localization
targets for visits affected by suspicious perturbations.
Knowledge-consistent plausible variations are further introduced
to distinguish reasonable medication changes from suspicious deviations.
To learn from these complementary supervision signals, we use shared
visit-level anomaly logits for both tasks, with account-level predictions
obtained through selective aggregation and the shared logits jointly
optimized by account-level and visit-level objectives.

\subsection{Problem Formulation}
\label{sec:problem_formulation}

For an account $a$, we represent its longitudinal medication trajectory as an ordered sequence of visits,
\begin{equation}
X_a =
\left(
V_{a,1}, V_{a,2}, \ldots, V_{a,T_a}
\right),
\end{equation}
where $V_{a,t}$ denotes the multiset of medication occurrences recorded at the $t$-th visit. Repeated medication occurrences within the same visit are retained. Since medications within a visit do not have a reliable internal ordering, $V_{a,t}$ is treated as an unordered multiset, whereas the longitudinal order across visits is preserved.

Given $X_a$, the model produces a local anomaly logit $u_{a,t}$ for each scoreable visit, indicating the strength of evidence that visit $t$ contains a suspicious medication deviation. The first visit is retained as historical context but is excluded from anomaly scoring. Accordingly, local anomaly logits are defined over $t = 2, \ldots, T_a$.

Rather than learning separate predictors for account-level detection and visit-level localization, we use visit-level anomaly logits as the shared evidence representation for both tasks. The account-level anomaly logit $s_a$ is derived directly from these local logits:
\begin{equation}
s_a =
g\!\left(
u_{a,2}, \ldots, u_{a,T_a}
\right),
\end{equation}
where $g(\cdot)$ denotes a selective aggregation function that combines informative local anomaly evidence into an account-level decision. Thus, account-level detection and visit-level localization are linked through the same visit-level anomaly logits, rather than being formulated as independent prediction tasks.

The following subsections describe the construction of proxy
supervision and the learning framework that uses these signals
for local-to-global anomaly detection.

\subsection{Knowledge-Guided Proxy Anomaly Construction}
\label{sec:proxy_construction}

Reliable ground-truth anomaly annotations are unavailable at either
the account or visit level. A central challenge in constructing
proxy supervision is that medication changes should not be
equated with anomalies, as longitudinal medication trajectories
may undergo substantial changes while remaining clinically
plausible. We therefore construct proxy anomalies through
controlled perturbations of original medication trajectories,
distinguishing plausible variations from suspicious deviations
according to the constraints imposed during candidate selection.

Specifically, we consider three perturbation mechanisms at
different structural granularities: K1 (\textit{Sparse
Replacement}), K2 (\textit{Semantic Episode Replacement}),
and K3 (\textit{Segment Injection}). K1 and K2 each include
Same and Cross variants, while K3 introduces trajectory-level
contextual perturbations. K1-Same and K2-Same represent
knowledge-consistent plausible variations, whereas
K1-Cross, K2-Cross, and K3 are treated as suspicious proxies.

Let $\mathcal{Q}$ denote the set of proxy variants:
\begin{equation}
\mathcal{Q} =
\{
\mathrm{K1\text{-}Same},
\mathrm{K1\text{-}Cross},
\mathrm{K2\text{-}Same},
\mathrm{K2\text{-}Cross},
\mathrm{K3}
\}.
\end{equation}

To provide a unified formulation of the perturbation
mechanisms and their variants, we characterize each proxy
variant $q$ by a mechanism configuration:
\begin{equation}
\Theta_q =
\left(
\tau_q^{\mathrm{tar}},
\tau_q^{\mathrm{rep}},
\Phi_q
\right),
\end{equation}
where $\tau_q^{\mathrm{tar}}$ and $\tau_q^{\mathrm{rep}}$
specify the types of target and replacement units,
respectively, and $\Phi_q$ denotes the corresponding
candidate-selection constraints. Given an original
trajectory $X_a$, the controlled perturbation operator
$\operatorname{P}$, parameterized by $\Theta_q$, produces
a perturbed trajectory and its associated affected-visit set:
\begin{equation}
\left(
\widetilde X_a^{(q)},
\mathcal I_a^{(q)}
\right)
=
\operatorname{P}
\left(
X_a;\Theta_q
\right).
\end{equation}
Here, $\widetilde X_a^{(q)}$ denotes the perturbed trajectory,
and $\mathcal I_a^{(q)}$ identifies the scoreable visits
directly affected by the perturbation.

\paragraph{Medication Knowledge Grounding}

We link medication records to corresponding drug entities in a
reliable drug knowledge graph (Drug KG). The resulting drug-group
relationships support candidate selection for K1-Same, while the
hierarchical therapeutic classification in the 2023 Chinese
National Reimbursement Drug Catalogue provides Level-1 and
Level-2 medication categories. Specifically, Level 1 is used
to define cross-category replacements in K1-Cross and K2-Cross
and to characterize medication segments in K3, while Level 2
defines therapeutic relatedness for K2-Same. We further use a
pretrained medication embedding function $E_{\mathrm{med}}(\cdot)$
to assess semantic similarity during candidate selection across
K1, K2, and K3. 
Knowledge-based constraints are applied to reliably mapped medications, while episode- and segment-level candidates must satisfy the required mapping-coverage criteria.

Based on the unified formulation and the above knowledge
resources, Table~\ref{tab:proxy_operators} summarizes the
three perturbation mechanisms in terms of their granularity,
replacement sources, core constraints, and the types of
deviations they introduce.

\begin{table*}[t]
\centering
\small
\renewcommand{\arraystretch}{1.35}
\setlength{\tabcolsep}{8pt}
\caption{Overview of the three proxy perturbation mechanisms.}
\label{tab:proxy_operators}
\begin{tabular}{
p{0.17\textwidth}
p{0.22\textwidth}
p{0.27\textwidth}
p{0.22\textwidth}}
\hline
\textbf{Component}
&
\textbf{K1: Sparse Replacement}
&
\textbf{K2: Semantic Episode Replacement}
&
\textbf{K3: Segment Injection}
\\
\hline

Perturbation granularity
&
Medication occurrence
&
Treatment episode
&
Contiguous visit segment
\\

Replacement source
&
Drug KG-grounded medication candidate
&
Real treatment episode
&
Real segment from another account
\\

Core constraints
&
Medication semantics
&
Treatment semantics and structural matching
&
Cross-account and segment constraints
\\

Introduced deviation
&
Medication-level semantic variation
&
Treatment-process-level variation
&
Trajectory-level contextual inconsistency
\\

\hline
\end{tabular}
\end{table*}

\paragraph{K1: Sparse Replacement.}

K1 introduces sparse medication-level perturbations by
replacing individual medication occurrences within visits.
For both K1-Same and K1-Cross, the target unit
$\tau_q^{\mathrm{tar}}$ is a medication occurrence selected
from a recipient visit, whereas the replacement unit
$\tau_q^{\mathrm{rep}}$ is a drug entity selected from
the Drug KG-grounded candidate pool. The candidate-selection
constraints $\Phi_q$ distinguish the two variants. For
K1-Same, the replacement drug must belong to the same
Drug KG group as the target medication. For K1-Cross,
it must belong to a different Level-1 therapeutic category.
Both variants exclude drugs already present in the recipient visit and restrict candidates to those with similar log-transformed occurrence frequencies in the training split. Within this frequency-restricted candidate pool, the replacement with the highest or lowest cosine similarity in the pretrained medication embedding space is selected for K1-Same or K1-Cross, respectively.

Under these variant-specific configurations, the
perturbation operator $\operatorname{P}$ replaces the
selected medication occurrence with the selected
replacement drug. Consequently, the affected-visit set
$\mathcal I_a^{(q)}$ identifies the recipient visit
containing the replaced occurrence.

\paragraph{K2: Semantic Episode Replacement.}

K2 extends medication-level perturbations to coherent
treatment episodes. The target unit $\tau_q^{\mathrm{tar}}$
is a treatment episode extracted from the recipient
trajectory, while the replacement unit
$\tau_q^{\mathrm{rep}}$ is a real treatment episode
selected from another account in the same city and
data split. We extract episodes spanning 7--30 days
and characterize each episode by its dominant
Level-1 and Level-2 therapeutic categories, determined
from the occurrence-weighted distribution of formally
mapped medications, with repeated occurrences counted
separately.

The candidate-selection constraints $\Phi_q$ incorporate
both therapeutic relationships and structural matching.
Candidate episodes are matched to the recipient episode
in visit count, temporal span, and medication volume.
For K2-Same, the replacement episode must share the
recipient episode's dominant Level-2 therapeutic
category and satisfy a minimum cosine-similarity
requirement in the pretrained medication embedding
space. Among eligible candidates, the episode with
the highest cosine similarity to the recipient episode
is selected, representing a therapeutically related
treatment alternative. For K2-Cross, the replacement
episode must have a different dominant Level-1
therapeutic category from the recipient episode,
while overlap in non-dominant categories is allowed.
Among structurally eligible candidates, the episode
with the lowest cosine similarity to the recipient
episode is selected.

Each episode is represented by the occurrence-weighted
mean of its medication embeddings:
\begin{equation}
\mathbf{p}(E)
=
\operatorname{Normalize}
\left(
\frac{1}{N_E}
\sum_{m\in E} E_{\mathrm{med}}(m)
\right),
\end{equation}
where $N_E$ denotes the total number of medication
occurrences in episode $E$, with repeated occurrences
counted separately. Cosine similarity between episode
representations is used to guide candidate selection
for both K2-Same and K2-Cross.

Under these variant-specific configurations, the
perturbation operator $\operatorname{P}$ replaces the
recipient episode with the selected real treatment
episode. Accordingly, the affected-visit set
$\mathcal I_a^{(q)}$ identifies the visits belonging
to the replaced episode. The 7--30-day duration
constraint is used only for proxy episode construction
and is not supplied as an elapsed-time feature to
the anomaly model.

\paragraph{K3: Segment Injection.}

K3 introduces trajectory-level contextual perturbations by
replacing a contiguous visit segment in the recipient
trajectory with a real segment from another account.
The target unit $\tau_q^{\mathrm{tar}}$ is a contiguous
segment selected from the recipient trajectory, whereas
the replacement unit $\tau_q^{\mathrm{rep}}$ is a real
segment extracted from a donor account. Both segments
have the same length $L$, where $3\leq L\leq 10$.

The candidate-selection constraints $\Phi_q$ require
the donor account to differ from the recipient account
while belonging to the same city and data split.
Candidate segments must satisfy medication-mapping coverage and structural matching constraints, share the same dominant broad drug group, and differ in their dominant Level-1 therapeutic category. Among eligible candidates, we select the segment with the lowest cosine similarity to the recipient segment in the pretrained medication embedding space.

Unlike K1 and K2, K3 does not replace individual
medications or treatment episodes. Instead, it introduces
contextual inconsistency by inserting a real segment
from another account into the recipient trajectory.
Because the donor segment comes from an actual trajectory,
its medication co-occurrence patterns and internal visit
order are preserved. The perturbation operator
$\operatorname{P}$ replaces the recipient segment with
the selected donor segment, and the affected-visit set
$\mathcal I_a^{(q)}$ identifies the recipient visits
belonging to the replaced segment.

\subsection{Local-to-Global Anomaly Scoring}
\label{sec:visit_scoring}

Our framework derives account-level anomaly predictions
from shared visit-level anomaly evidence. Given a
longitudinal medication trajectory, the model first
encodes the medication composition of each visit into
a fixed-dimensional representation. These visit
representations are then processed in their longitudinal
order by a causal visit-sequence encoder to capture
dependencies across visits. A shared anomaly scoring
function subsequently produces a local anomaly logit
for each scoreable visit. Finally, the account-level
anomaly logit is obtained through selective aggregation
of these local scores. The resulting visit-level logits
serve as shared evidence for both account-level detection
and visit-level localization.

\paragraph{Medication-Multiset Encoding}
For each medication occurrence $m_{a,t,j}$ in visit
$V_{a,t}$, we first obtain a pretrained medication
embedding:
\begin{equation}
\mathbf{e}_{a,t,j}
=
E_{\mathrm{med}}(m_{a,t,j})
\in \mathbb{R}^{d_1},
\end{equation}
where $E_{\mathrm{med}}(\cdot)$ denotes the pretrained
medication embedding function introduced above. The
embedding lookup is kept frozen during anomaly training.
A trainable linear projection followed by layer
normalization maps each medication embedding into a
shared hidden representation space:
\begin{equation}
\mathbf{z}_{a,t,j}
=
\operatorname{LN}
\left(
\mathbf{W}_{p}\mathbf{e}_{a,t,j}
+
\mathbf{b}_{p}
\right)
\in \mathbb{R}^{d_2},
\end{equation}
where $\mathbf{W}_{p}\in\mathbb{R}^{d_2\times d_1}$
and $\mathbf{b}_{p}\in\mathbb{R}^{d_2}$ are trainable
parameters. The projection and normalization
parameters are optimized during anomaly training.
Padding positions are assigned zero representations
and excluded from attention through a key-padding mask.

Specifically, we introduce a learnable
\texttt{[VISIT]} token
$\mathbf{c}_{\mathrm{VISIT}}\in\mathbb{R}^{d_2}$
and prepend it to the projected medication
representations. The resulting sequence is processed
by a Transformer encoder, and the output corresponding
to the \texttt{[VISIT]} token is taken as the visit
representation:
\begin{equation}
\mathbf{v}_{a,t}
=
\operatorname{VisitEnc}
\left(
[\mathbf{c}_{\mathrm{VISIT}};
\mathbf{z}_{a,t,1};
\ldots;
\mathbf{z}_{a,t,n_{a,t}}]
\right)_{[\mathrm{VISIT}]},
\end{equation}
where $\mathbf{z}_{a,t,j}$ denotes the projected
representation of medication occurrence $m_{a,t,j}$,
and $n_{a,t}$ is the number of medication occurrences
in the visit. $\operatorname{VisitEnc}$ denotes the
Transformer encoder followed by a final layer
normalization.

No positional encoding is introduced among medication
tokens. Consequently, the resulting visit representation
is permutation-invariant to the ordering of medication
occurrences. Repeated occurrences are retained rather
than deduplicated, allowing the representation to reflect
medication multiplicity.

\paragraph{Causal Visit-Sequence Modeling}
The visit representations are arranged in their original
longitudinal order and processed by a causal visit-sequence
Transformer to capture dependencies across visits. To
represent relative visit positions, we introduce a learnable
attention bias. For attention head $r$, the bias between
visits $i$ and $j$ is defined as
\begin{equation}
b_{ij}^{(r)}
=
B^{(r)}
\left[
\operatorname{clip}(j-i,-R,R)+R
\right],
\end{equation}
where $B^{(r)}$ is a head-specific learnable bias table
and $R$ denotes the maximum relative-position range.

The attention logits combine scaled dot-product attention
with the relative-position bias and a causal attention mask:
\begin{equation}
\ell_{ij}^{(r)}
=
\frac{
(\mathbf{q}_{i}^{(r)})^\top
\mathbf{k}_{j}^{(r)}
}{
\sqrt{d_2/H}
}
+
b_{ij}^{(r)}
+
M_{ij},
\end{equation}
where $\mathbf{q}_{i}^{(r)}$ and $\mathbf{k}_{j}^{(r)}$
are the query and key representations for head $r$,
$H$ is the number of attention heads, and $M_{ij}$
masks future visits ($j>i$) and padded positions.
Consequently, each visit can attend only to itself
and preceding visits.

No absolute positional embeddings or elapsed-time
embeddings are introduced. Instead, longitudinal
dependencies are captured through causal attention
and learned relative visit-position biases, without
explicitly modeling actual time intervals between visits.

The resulting contextualized visit representations are
denoted by
\begin{equation}
(\mathbf{h}_{a,1},\ldots,\mathbf{h}_{a,T_a})
=
\operatorname{VisitSeqEnc}
(\mathbf{v}_{a,1},\ldots,\mathbf{v}_{a,T_a}).
\end{equation}
Due to causal attention, each $\mathbf{h}_{a,t}$
incorporates the medication representation of the
current visit and the available history
$\mathbf{v}_{a,1:t}$. The first visit is therefore
retained as historical context but excluded from
anomaly scoring.

Both Transformer encoders (VisitEnc and VisitSeqEnc) use four attention heads,
a hidden dimension $d_2$, and a feed-forward dimension
$d_{\mathrm{ff}}$, with pre-layer normalization,
GELU activation, and dropout of 0.1, followed by
a final layer normalization. The Visit Encoder consists
of one Transformer layer, whereas the Visit-Sequence
Encoder consists of two layers.

\paragraph{Shared Visit-Level Anomaly Scoring}

For each scoreable visit $t\geq 2$, a shared Visit
Anomaly Head maps its contextualized representation
$\mathbf{h}_{a,t}$ to a scalar anomaly logit:
\begin{equation}
u_{a,t}
=
f_{\phi}(\mathbf{h}_{a,t}),
\end{equation}
where $f_{\phi}$ is a two-layer MLP with GELU
activation and dropout of 0.1 between its linear
layers. Larger logits indicate stronger local evidence
of a suspicious medication deviation.

The same scoring function is shared across K1, K2,
and K3, without introducing operator-specific
prediction heads or providing operator identity
as an input. This allows heterogeneous proxy
perturbations to be evaluated through a common
visit-level anomaly scoring mechanism. The resulting
logits serve as the local evidence for visit-level
localization and are subsequently aggregated to
derive account-level anomaly predictions.

\paragraph{Top-$\alpha$ Local-to-Global Aggregation.}

The account-level anomaly logit is derived by selectively
aggregating the shared visit-level anomaly logits. Let
\begin{equation}
D_a = T_a - 1
\end{equation}
denote the number of scoreable visits, and define the
number of retained logits as
\begin{equation}
k_a
=
\max
\left(
1,
\left\lceil \alpha D_a \right\rceil
\right),
\end{equation}
where $\alpha$ controls the proportion of local anomaly
evidence retained for aggregation. Let
$\mathbf{u}_a=(u_{a,2},\ldots,u_{a,T_a})$, and let
$\operatorname{Top}_{k_a}(\mathbf{u}_a)$ denote the
$k_a$ largest logits in this sequence. The account-level
anomaly logit is then computed as
\begin{equation}
s_a
=
\frac{1}{k_a}
\sum_{u\in\operatorname{Top}_{k_a}(\mathbf{u}_a)}u.
\end{equation}

This aggregation is motivated by the observation that
suspicious medication deviations may be concentrated
in a limited subset of visits rather than distributed
uniformly across an entire trajectory. Averaging over
all visits may dilute informative local evidence with
unaffected visits, whereas relying solely on the maximum
score makes the account-level prediction sensitive
to a single potentially noisy visit. Top-$\alpha$
aggregation provides an intermediate strategy by
combining multiple high-scoring visits while limiting
the contribution of less informative ones. Its
proportional selection also accommodates trajectories
of different lengths.

Importantly, aggregation is performed directly on
the raw visit-level logits, before applying the sigmoid
function. The resulting account-level logit $s_a$
is subsequently transformed into a probability for
account-level classification. Thus, the account-level
prediction is derived from the same local anomaly
evidence used for visit-level localization.

\subsection{Joint Account- and Visit-Level Learning}
\label{sec:joint_learning}

The objective of joint learning is to make local anomaly
evidence support both account-level detection and
visit-level localization. Account-level labels indicate
whether a trajectory contains suspicious deviations,
while affected-visit targets identify where these
deviations occur. We combine these complementary
supervision signals to jointly constrain the same
visit-level anomaly logits, from which account-level
predictions are derived through Top-$\alpha$ aggregation.

Since training includes both original trajectories and
their perturbed variants, we define the set of training
instance types as
\begin{equation}
\mathcal R
=
\{\mathrm{Original}\}\cup\mathcal Q,
\end{equation}
where $r\in\mathcal R$ indexes the six training-instance
types. Let $X_a^{(r)}$ denote the corresponding trajectory,
which is either the original trajectory or a perturbed
trajectory. The model produces visit-level logits
$u_{a,t}^{(r)}$ and an account-level logit $s_a^{(r)}$.

Each training instance receives a binary account-level
label:
\begin{equation}
y_a^{(r)}
=
\begin{cases}
0, & r\in\{\mathrm{Original},
\mathrm{K1\text{-}Same},
\mathrm{K2\text{-}Same}\},\\
1, & r\in\{\mathrm{K1\text{-}Cross},
\mathrm{K2\text{-}Cross},
\mathrm{K3}\}.
\end{cases}
\end{equation}
The account-level objective is
\begin{equation}
\mathcal L_{\mathrm{acc}}
=
\operatorname{BCE}
\left(
\sigma(s_a^{(r)}),y_a^{(r)}
\right).
\end{equation}
Since the account-level logit is derived through
Top-$\alpha$ aggregation of visit-level logits, this
objective directly supervises the local evidence
contributing to the global prediction.

For suspicious proxy variants, the affected-visit sets
$\mathcal I_a^{(q)}$ provide positive localization
targets. The remaining scoreable visits form the
unaffected set:
\begin{equation}
\mathcal U_a^{(q)}
=
\{2,\ldots,T_a\}\setminus\mathcal I_a^{(q)}.
\end{equation}

To prevent the larger unaffected group from dominating
localization supervision, we adopt class-balanced
binary cross-entropy, assigning equal total weight
to affected and unaffected visits. For each suspicious
proxy, the localization loss is
\begin{equation}
\begin{aligned}
\ell_{\mathrm{loc}}^{(a,q)}
=
\frac{1}{2}
\Bigg[
&\frac{1}{|\mathcal I_a^{(q)}|}
\sum_{t\in\mathcal I_a^{(q)}}
\operatorname{softplus}(-u_{a,t}^{(q)})
\\
+&\frac{1}{|\mathcal U_a^{(q)}|}
\sum_{t\in\mathcal U_a^{(q)}}
\operatorname{softplus}(u_{a,t}^{(q)})
\Bigg],
\end{aligned}
\end{equation}
where $\operatorname{softplus}(x)=\log(1+\exp(x))$.
If no unaffected scoreable visits exist, the positive-class mean alone is used.
The per-proxy losses are
averaged equally across suspicious proxy instances
in the minibatch to obtain $\mathcal L_{\mathrm{loc}}$.
This supervision applies only to K1-Cross, K2-Cross,
and K3; Original, K1-Same, and K2-Same receive no
localization supervision.

The final objective combines the two supervision signals:
\begin{equation}
\mathcal L
=
\mathcal L_{\mathrm{acc}}
+
\lambda_{\mathrm{local}}\mathcal L_{\mathrm{loc}},
\end{equation}
where $\lambda_{\mathrm{local}}$ controls the contribution
of localization supervision. Setting
$\lambda_{\mathrm{local}}=0$ yields L2G-AccountOnly,
which uses account-level supervision alone, whereas
$\lambda_{\mathrm{local}}>0$ yields L2G-Joint, which
jointly optimizes account-level detection and visit-level
localization through the shared visit-level anomaly logits.

\section{Experiment}

\subsection{Setup}

We evaluate all methods on longitudinal medication trajectories collected
from two independent cities, denoted City-A and City-B. Each account is
represented as a temporally ordered sequence of medication visits, where
each visit contains the set of medications recorded for that encounter.
To ensure that temporal modeling and visit-level localization are meaningful,
we retain only accounts with at least two medication visits.

For City-A, 8,855 eligible accounts are partitioned at the account level
into 6,199 training, 885 validation, and 1,771 test accounts using a
70/10/20 split with seed 42, ensuring that no account appears in more than
one split. City-B contains 562 eligible accounts and is evaluated using a
burden-controlled five-fold out-of-fold protocol because its substantially
smaller cohort and shorter trajectories make an additional fixed validation
split unreliable for some proxy operators. Each City-B account serves as
held-out data exactly once, and all training-dependent preprocessing and
proxy construction are performed independently within each fold.

Following the knowledge-guided proxy construction described in
Section~\ref{sec:proxy_construction}, Original, K1-Same, and K2-Same
trajectories are treated as plausible negatives, whereas K1-Cross,
K2-Cross, and K3 constitute suspicious proxies.
Table~\ref{tab:benchmark_composition} summarizes the resulting City-A
benchmark composition. Because the three proxy operators have different
applicability constraints, the numbers of successfully generated trajectories
vary across operators.

\begin{table*}[t]
\centering
\caption{Overall performance on the City-A test set. Macro metrics are
unweighted averages across K1, K2, and K3. Best results are shown in
\textbf{bold} and second-best results are \underline{underlined}; rankings
are determined using unrounded scores. ``--'' indicates that a method
does not provide a native visit-level anomaly score.}
\label{tab:overall_results}

\small
\setlength{\tabcolsep}{2.4pt}
\renewcommand{\arraystretch}{0.98}

\begin{tabular*}{\textwidth}{
@{\extracolsep{\fill}}
l
rrrrrr
@{}
}
\toprule
Method
& \shortstack{K1 AP}
& \shortstack{K2 AP}
& \shortstack{K3 AP}
& \shortstack{Macro AP}
& \shortstack{Macro AUROC}
& \shortstack{Loc. Macro AP} \\
\midrule

\multicolumn{7}{@{}l}{\textit{Native anomaly methods}} \\
Deep SVDD-Native
& 0.5070 & 0.5130 & 0.5184 & 0.5128 & 0.5208 & 0.3462 \\
NCAD-Native
& 0.5017 & 0.5523 & 0.5472 & 0.5338 & 0.5334 & 0.3501 \\
AnomalyBERT-Native
& 0.5091 & 0.5839 & 0.5628 & 0.5519 & 0.5606 & 0.3745 \\
TranAD-Native
& 0.4969 & 0.5231 & 0.5479 & 0.5226 & 0.5352 & 0.3528 \\
CAT-Native
& 0.5000 & 0.5000 & 0.5000 & 0.5000 & 0.5000 & 0.3689 \\
OC4Seq-Native
& 0.4978 & 0.5689 & 0.5839 & 0.5502 & 0.5585 & 0.4019 \\
EHR-BERT-Native
& 0.5054 & 0.5214 & 0.5169 & 0.5145 & 0.5247 & 0.4381 \\
EHR-HOTN-Native
& 0.5005 & 0.5063 & 0.5058 & 0.5042 & 0.5071 & 0.2211 \\

\addlinespace[2pt]
\multicolumn{7}{@{}l}{\textit{Proxy-adapted literature methods}} \\
Deep SAD-Proxy
& \underline{0.7914}
& 0.7360
& 0.7110
& 0.7461
& 0.7304
& -- \\
TAIL-MIL-Proxy
& 0.5248
& 0.5981
& 0.5986
& 0.5738
& 0.5910
& 0.5153 \\
AnomalyBERT-Proxy
& 0.5804
& \underline{0.7847}
& 0.7514
& 0.7055
& 0.6806
& \underline{0.6605} \\
NCAD-Proxy
& 0.6290
& 0.7782
& \underline{0.7657}
& 0.7243
& 0.7137
& 0.5706 \\

\addlinespace[2pt]
\multicolumn{7}{@{}l}{\textit{Controlled models and ours}} \\
Account-Pooled Transformer
& 0.7821
& 0.7708
& 0.7449
& \underline{0.7659}
& \underline{0.7439}
& -- \\
L2G-AccountOnly
& 0.7323
& 0.7536
& 0.7296
& 0.7385
& 0.7280
& 0.6380 \\
\textbf{L2G-Joint}
& \textbf{0.7981}
& \textbf{0.7882}
& \textbf{0.7669}
& \textbf{0.7844}
& \textbf{0.7707}
& \textbf{0.7215} \\

\bottomrule
\end{tabular*}
\end{table*}

Specifically, for \emph{account-level anomaly detection}, we construct three separate matched evaluation sets for K1, K2, and K3. Each matched pair consists of an original trajectory and its corresponding suspicious proxy generated from the same account, yielding Original--K1-Cross, Original--K2-Cross, and Original--K3 comparisons. In City-A, the validation sets contain 871, 780, and 632 matched pairs for K1, K2, and K3, respectively, while the corresponding test sets contain 1,736, 1,565, and 1,281 pairs. For each operator, we report average precision (AP) and area under the receiver operating characteristic curve (AUROC), and compute Operator-Macro AP and Operator-Macro AUROC as the unweighted averages over K1, K2, and K3.
For \emph{visit-level anomaly localization}, visits directly modified during proxy construction are treated as positive locations, and localization AP is computed from model-produced visit-level anomaly scores over scoreable visits in suspicious trajectories. Localization AP is computed separately for K1, K2, and K3, with their unweighted average reported as Localization Macro AP. The first visit serves exclusively as historical context and is therefore excluded from localization scoring. For methods that natively produce visit-level anomaly scores but do not define an account-level anomaly score, we use the mean of the highest-scoring 20\% of scoreable visits as the account-level score; this aggregation is used only for evaluation and does not participate in model training. 
For City-A, checkpoint selection for all trainable methods is based solely
on validation Operator-Macro AP, and the test set is evaluated only once
after model selection and checkpoint freezing. The City-B evaluation follows
the five-fold OOF protocol described in Section~\ref{sec:cityb} and does not
use City-B held-out performance for model selection.

We organize the comparison methods into three groups. 
\emph{Native anomaly methods} retain their original anomaly-learning formulations and use untouched training trajectories as the underlying data source, including Deep SVDD~\citep{ruff2018deepsvdd}, NCAD~\citep{carmona2022ncad}, AnomalyBERT~\citep{jeong2023anomalybert}, TranAD~\citep{tuli2022tranad}, CAT~\citep{zhang2022cat}, OC4Seq \citep{wang2021oc4seq}, EHR-BERT~\citep{niu2024ehrbert}, and EHR-HOTN~\citep{niu2022anomalous}. 
\emph{Proxy-adapted literature methods} replace their original anomaly source with the proposed K1/K2/K3 proxies while otherwise preserving the characteristic learning formulation; this group includes Deep SAD~\citep{ruff2020deepsad}, TAIL-MIL~\citep{jang2025tailmil}, AnomalyBERT~\citep{jeong2023anomalybert}, and NCAD~\citep{carmona2022ncad}. 
Finally, the controlled models include \emph{Account-Pooled Transformer}, which predicts directly from a pooled trajectory representation, \emph{L2G-AccountOnly}, which uses visit-level anomaly scores and Top-$\alpha$ aggregation with account-level supervision only, and the proposed \emph{L2G-Joint}, which further incorporates visit-level localization supervision. Localization results are reported only for methods that naturally provide visit-level anomaly scores.

All experiments use a fixed random seed of 42. City-A uses the fixed
account-level split described above, whereas City-B uses the common five-fold
OOF partition shared by all comparison methods. For City-A, medication
identifiers are mapped to a Rare-20 vocabulary containing 1,697 tokens.
Baseline methods operating on continuous visit representations use the same
frozen 512-dimensional local VisitEncoder when compatible with their original
formulations, whereas methods with native token- or event-level
representations, such as EHR-BERT and EHR-HOTN, retain their method-specific
input representations. For our models, PULSE-derived medication embeddings
are kept frozen, while the visit encoder and temporal sequence model are
optimized on the training split. For the proposed L2G models, $\alpha$ and
$\lambda_{\mathrm{local}}$ are selected through training-consistent grid
searches on the City-A validation set using Operator-Macro AP, yielding
$\alpha=0.20$ and $\lambda_{\mathrm{local}}=2.0$. These settings are then
fixed for all subsequent evaluations. Method-specific architectures,
optimization settings, and task-specific adaptations are provided in the
Appendix.

\subsection{Overall Performance}

Table~\ref{tab:overall_results} summarizes the overall performance on City-A.
L2G-Joint achieves the highest value in every metric column of
Table~\ref{tab:overall_results}, with an
Operator-Macro AP of 0.7844, an Operator-Macro AUROC of 0.7707, and a
Localization Macro AP of 0.7215. The strongest competing account-level model,
Account-Pooled Transformer, reaches a Macro AP of 0.7659 and a Macro AUROC of
0.7439, while the strongest competing localization result is obtained by
AnomalyBERT-Proxy with a Localization Macro AP of 0.6605. Relative to these
strongest competing results, L2G-Joint improves Macro AP, Macro AUROC, and
Localization Macro AP by absolute margins of 0.0185, 0.0268, and 0.0610,
corresponding to relative gains of approximately 2.4\%, 3.6\%, and 9.2\%,
respectively. These results show that L2G-Joint provides consistent gains in
both account-level anomaly detection and visit-level anomaly localization.

A clear performance gap is also visible across the three groups of comparison
methods. Native anomaly methods remain within a relatively narrow Macro-AP
range of 0.5000--0.5519, despite covering diverse anomaly-learning paradigms.
Proxy-adapted literature methods perform substantially better, ranging from
0.5738 to 0.7461 Macro AP. This pattern suggests that performance depends not
only on model architecture, but also strongly on whether the anomaly
supervision is aligned with the medication deviations targeted by the
benchmark. We examine this effect more directly in the following subsection.

\begin{figure*}[t]
    \centering
    \includegraphics[width=1.4\columnwidth]{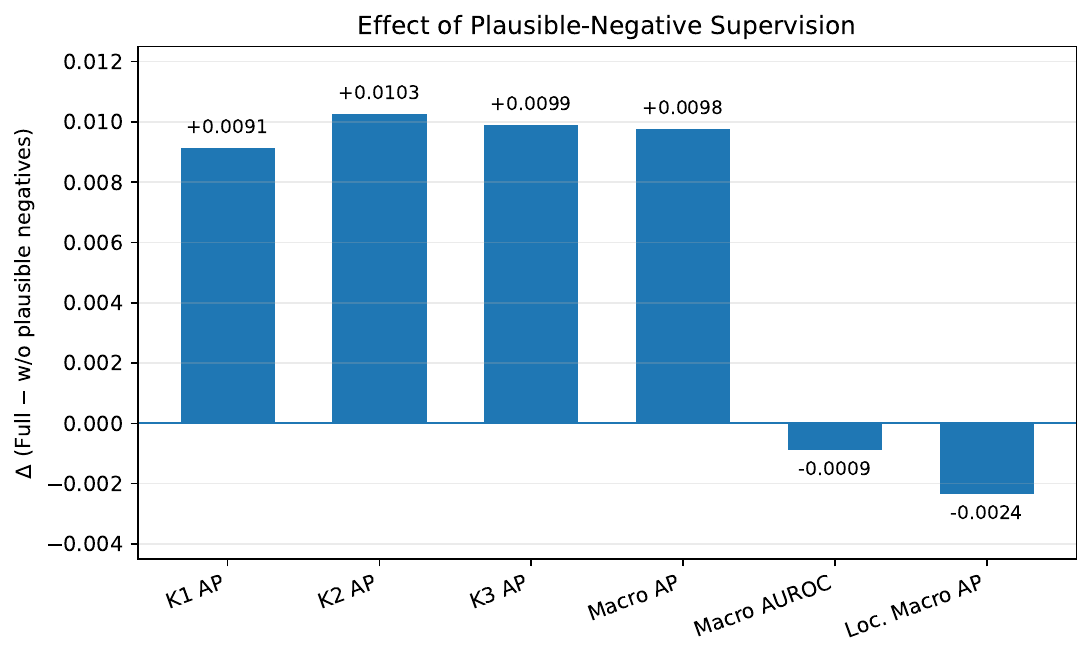}
    \caption{Effect of plausible-negative supervision on City-A.
    Bars show the metric difference between the full L2G-Joint model
    and the ablation trained without K1-Same and K2-Same.
    Plausible negatives consistently improve operator-specific and Macro AP,
    while producing negligible changes in Macro AUROC and Localization Macro AP.}
    \label{fig:plausible_negative_ablation}
\end{figure*}

The controlled models further clarify the role of local anomaly evidence and
visit-level supervision. L2G-AccountOnly, which uses the same visit-level
scoring architecture as the full model but is trained only with account-level
supervision, achieves a Macro AP of 0.7385 and Macro AUROC of 0.7280.
Account-Pooled Transformer performs better at the account level, reaching
0.7659/0.7439, indicating that forcing the global decision through visit-level
anomaly scores is not by itself sufficient to improve detection. With explicit
visit-level supervision, however, L2G-Joint increases Macro AP to 0.7844 and
Macro AUROC to 0.7707, while Localization Macro AP rises from 0.6380 for
L2G-AccountOnly to 0.7215. These results support the role of localization
supervision in reducing ambiguity about which local deviations should support
the global account-level decision.

The operator-specific results show a similar pattern of consistently strong
performance across anomaly scales. L2G-Joint achieves the highest AP on K1,
K2, and K3, with scores of 0.7981, 0.7882, and 0.7669, respectively.
However, the margins over the strongest operator-specific competitors are
relatively small: Deep SAD-Proxy reaches 0.7914 on K1, AnomalyBERT-Proxy
reaches 0.7847 on K2, and NCAD-Proxy reaches 0.7657 on K3. The aggregate
advantage of L2G-Joint therefore does not arise from a disproportionate gain
on a single anomaly type, but from maintaining consistently strong performance
across sparse, episode-level, and segment-level deviations.

\subsection{Analysis of Knowledge-Guided Proxy Construction}

The overall results show large performance gains when existing anomaly
detectors are trained with the knowledge-guided proxy anomalies. We further
examine whether these gains reflect better alignment between the learned
anomaly notion and the target medication deviations, and whether the
plausible-negative component of the proxy construction contributes
independently to this alignment.

We first distinguish anomaly-definition mismatch from optimization failure
using AnomalyBERT-Native. After freezing its selected checkpoint, we evaluate
the model on the Soft, Uniform, and Peak degradations used by its native
self-supervised objective, without retraining or checkpoint reselection.
The model achieves a position-level Macro AP of 0.9900 and AUROC of 0.9984
on these native degradations, indicating that its original anomaly task is
learned almost perfectly. In contrast, the same frozen model achieves an
Operator-Macro AP of only 0.5519 and AUROC of 0.5606 on the knowledge-guided
K1--K3 benchmark. Although the two evaluations operate at different scoring
levels and are therefore not directly comparable as absolute metrics, the
contrast shows that the weak benchmark performance is not explained simply
by failure to learn the native objective. Together with the substantial
Native-to-Proxy gains observed for both AnomalyBERT and NCAD, this result
supports the importance of aligning anomaly supervision with the medication
deviations targeted by the benchmark.

We next assess the role of K1-Same and K2-Same as plausible negatives by
comparing the full L2G-Joint model with an otherwise matched ablation
that excludes both Same subtypes from training. As shown in
Fig.~\ref{fig:plausible_negative_ablation}, plausible-negative supervision
improves AP consistently across K1, K2, and K3 by 0.0091, 0.0103, and
0.0099, respectively, resulting in a Macro-AP gain of 0.0098. In contrast,
Macro AUROC changes by only $-0.0009$ and Localization Macro AP by
$-0.0024$. The benefit is therefore concentrated on account-level ranking
rather than localization or a uniform improvement across all metrics.

\begin{figure*}[t]
    \centering
    \includegraphics[width=1.3\columnwidth]{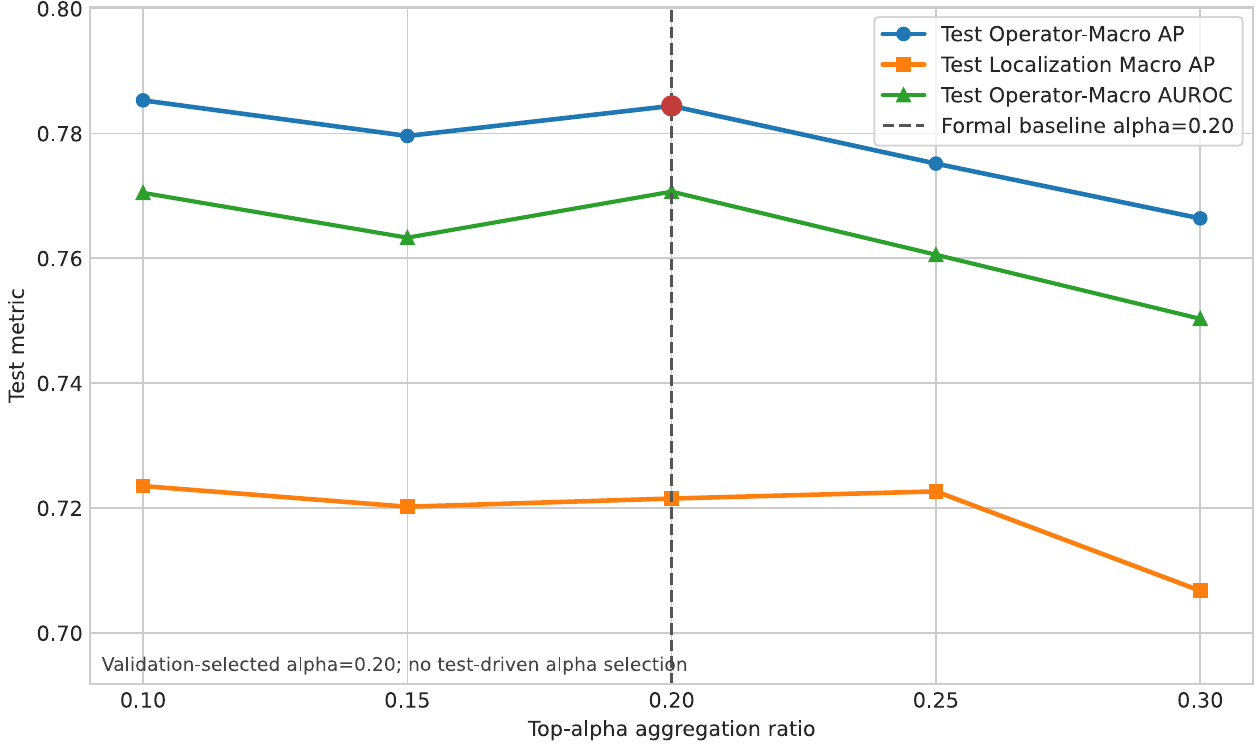}
    \caption{
    Training-consistent sensitivity to the Top-$\alpha$ aggregation ratio on
    City-A. Each point corresponds to an independently trained
    L2G-Joint model using the same $\alpha$ during training, validation,
    and testing. The validation-selected $\alpha=0.20$ is retained as the
    formal setting.
    }
    \label{fig:top_alpha_sensitivity}
\end{figure*}

\begin{table*}[t]
\centering
\caption{Training-consistent comparison of local-to-global aggregation
mechanisms on the City-A test set. Best results are shown in bold.}
\label{tab:aggregation_ablation}
\small
\setlength{\tabcolsep}{3.5pt}
\begin{tabular}{lcccccc}
\toprule
Aggregation &
K1 AP &
K2 AP &
K3 AP &
Macro AP &
Macro AUROC &
Loc. Macro AP \\
\midrule
Mean
& 0.7997
& 0.7698
& 0.7486
& 0.7727
& 0.7588
& 0.7257 \\
Max
& \textbf{0.8111}
& 0.7765
& 0.7470
& 0.7782
& 0.7628
& \textbf{0.7304} \\
Top-20\%
& 0.7981
& \textbf{0.7882}
& \textbf{0.7669}
& \textbf{0.7844}
& \textbf{0.7707}
& 0.7215 \\
\bottomrule
\end{tabular}
\end{table*}

The account-score distributions provide a descriptive view of how
plausible-negative supervision changes score behavior. Pooling K1-Same and
K2-Same as the Same group and their corresponding Cross variants as the
Cross group, both models preserve the ordering
Original $<$ Same $<$ Cross. With plausible-negative supervision, the mean
logits for Original, Same, and Cross are $-0.8045$, $-0.5386$, and $3.3771$,
respectively, compared with $-0.1419$, $0.2718$, and $2.5902$ without
plausible negatives. Because the two models are trained independently, the
absolute logit scales are not directly calibrated across runs. We therefore
interpret these score statistics descriptively, while the AP improvements
reported above provide the primary evidence that plausible negatives improve
account-level ranking.
Taken together, the matched ablation shows that plausible negatives improve
account-level ranking, while the score statistics descriptively illustrate
how they alter the placement of Original, Same, and Cross trajectories.
These findings support their role in defining the supervision boundary between
knowledge-consistent variation and suspicious deviation.

\subsection{Effect of Local-to-Global Aggregation}
\label{sec:aggregation_analysis}

We next examine how visit-level anomaly evidence should be aggregated into an
account-level decision. We first study the aggregation ratio within the
Top-$\alpha$ family and then compare the validation-selected Top-20\%
aggregation with full-sequence Mean and single-visit Max. Importantly, all
comparisons are training-consistent: the corresponding aggregation rule is
used identically during training, validation, and testing.

\paragraph{Selection and sensitivity of the Top-$\alpha$ ratio}
We independently train L2G-Joint with
$\alpha\in\{0.10,0.15,0.20,0.25,0.30\}$ while keeping all other settings
fixed. As shown in Fig.~\ref{fig:top_alpha_sensitivity}, $\alpha=0.20$
achieves the highest validation Operator-Macro AP (0.7740) and Macro AUROC
(0.7659), and is therefore retained as the formal aggregation ratio.
Performance remains relatively stable over $\alpha=0.10$--$0.20$, whereas
larger aggregation ratios lead to a clear decline in account-level detection.
In particular, increasing $\alpha$ to 0.30 reduces test Macro AP from 0.7844
to 0.7664 and Macro AUROC from 0.7707 to 0.7503. This trend suggests that
aggregating an overly large fraction of visits can dilute localized anomaly
evidence.

\paragraph{Comparison of aggregation mechanisms}
Having selected $\alpha=0.20$ on the validation set, we next compare
Top-20\% aggregation with two natural alternatives: Mean, which averages all
scoreable visit logits, and Max, which retains only the highest-scoring visit.
As summarized in Table~\ref{tab:aggregation_ablation}, Top-20\% achieves the
best overall account-level performance, with a Macro AP of 0.7844 and Macro
AUROC of 0.7707, compared with 0.7727/0.7588 for Mean and 0.7782/0.7628 for
Max.

\begin{table*}[t]
\centering
\caption{
Burden-controlled five-fold out-of-fold evaluation on City-B.
All methods use the same outer folds and held-out proxy samples, with model
configurations and training durations frozen from City-A.
Metrics are first computed within each held-out fold and then aggregated for
each operator using the number of matched account pairs as weights; raw scores
from independently trained folds are not pooled.
The evaluation contains 140, 43, and 121 held-out matched accounts for K1,
K2, and K3, respectively.
Best and second-best results are shown in bold and underlined, respectively.
}
\label{tab:cityb_oof}

\small
\setlength{\tabcolsep}{2.4pt}
\renewcommand{\arraystretch}{0.98}

\begin{tabular*}{\textwidth}{
@{\extracolsep{\fill}}
l
rrrrrr
@{}
}
\toprule
Method
& \shortstack{K1 AP}
& \shortstack{K2 AP}
& \shortstack{K3 AP}
& \shortstack{Macro\\AP}
& \shortstack{Macro\\AUROC}
& \shortstack{Loc. Macro\\AP} \\
\midrule

\multicolumn{7}{@{}l}{\textit{Native anomaly methods}} \\
Deep SVDD
& 0.5264 & 0.5321 & 0.5522 & 0.5369 & 0.5239 & 0.3207 \\
NCAD-Native
& 0.5479 & 0.6080 & \textbf{0.6763} & 0.6107 & 0.5863 & 0.3692 \\
AnomalyBERT-Native
& 0.5413 & 0.5558 & 0.5401 & 0.5457 & 0.5287 & 0.3931 \\
TranAD-Native
& 0.5331 & 0.5296 & 0.5236 & 0.5288 & 0.5125 & 0.3203 \\
CAT-Native
& 0.5098 & 0.5513 & 0.5132 & 0.5247 & 0.5049 & 0.3001 \\
OC4Seq-Native
& 0.5243 & 0.5381 & 0.5187 & 0.5270 & 0.4933 & 0.2730 \\
EHR-BERT-Native
& 0.5063 & 0.5247 & 0.5002 & 0.5104 & 0.4941 & 0.3152 \\
EHR-HOTN-Native
& 0.5240 & 0.5895 & 0.5510 & 0.5548 & 0.5332 & 0.1678 \\

\addlinespace[2pt]
\multicolumn{7}{@{}l}{\textit{Proxy-adapted literature methods}} \\
Deep SAD-Proxy
& 0.6349 & 0.6042 & 0.6242 & 0.6211 & 0.6166 & -- \\
TAIL-MIL-Proxy
& 0.5509 & 0.6160 & 0.5498 & 0.5722 & 0.5406 & 0.3199 \\
AnomalyBERT-Proxy
& 0.5867 & \textbf{0.6597} & 0.6325 & 0.6263 & 0.6136
& \underline{0.4657} \\
NCAD-Proxy
& 0.6055 & \underline{0.6300} & \underline{0.6754}
& 0.6370 & 0.6075 & 0.3664 \\

\addlinespace[2pt]
\multicolumn{7}{@{}l}{\textit{Controlled models and ours}} \\
Account-Pooled Transformer
& 0.6441 & 0.5620 & 0.6173 & 0.6078 & 0.5749 & -- \\
L2G-AccountOnly
& \underline{0.6756} & 0.6296 & 0.6530
& \textbf{0.6527} & \textbf{0.6287} & 0.3637 \\
\textbf{L2G-Joint}
& \textbf{0.6847} & 0.6006 & 0.6263
& \underline{0.6372} & \underline{0.6285} & \textbf{0.4985} \\

\bottomrule
\end{tabular*}
\end{table*}

The operator-specific results in Table~\ref{tab:aggregation_ablation} reveal
a clear scale-dependent pattern. Max performs best on K1, where the deviation
is sparse and can often be represented by a single strong local signal. In
contrast, Top-20\% performs best on K2 and K3, where anomalous evidence spans
treatment episodes or contiguous segments and is better represented by
multiple supporting visits. Mean is consistently weaker than Top-20\% for
account-level detection, consistent with localized anomaly evidence being
diluted by unaffected positions. Max yields the highest Localization Macro AP,
suggesting a modest trade-off between sharper localization and more robust
account-level detection.

Paired account-level bootstrap further supports these observations. Relative
to Mean, Top-20\% improves Macro AP by 0.0117
(95\% CI $[0.0039,0.0197]$) and Macro AUROC by 0.0118
(95\% CI $[0.0050,0.0188]$). Relative to Max, Top-20\% improves Macro AUROC
by 0.0078 (95\% CI $[0.0011,0.0147]$), whereas the Macro-AP difference of
0.0062 has a confidence interval that includes zero
(95\% CI $[-0.0021,0.0145]$).

Taken together, Fig.~\ref{fig:top_alpha_sensitivity} and
Table~\ref{tab:aggregation_ablation} support the selective local-to-global
aggregation design: a limited set of high-scoring visits provides a more
robust account-level representation than averaging the entire trajectory,
while avoiding the dependence on a single extreme visit introduced by Max.

\subsection{Cross-City Robustness on City-B}
\label{sec:cityb}

City-B contains a smaller cohort of 562 eligible accounts with shorter
medication trajectories. We therefore use burden-controlled five-fold
out-of-fold (OOF) evaluation to make efficient use of the limited cohort while
preventing local perturbations from affecting an excessively large fraction
of short trajectories. Each account serves as held-out data exactly once.
Within each fold, training-dependent assets are constructed and models are
trained using only the corresponding training accounts, while proxy donors
are restricted to the same data partition as the recipient. Metrics are first
aggregated across folds within each operator using matched-pair counts as
weights and then macro-averaged equally over K1--K3; raw scores from different
fold-specific models are not pooled. Burden constraints, training settings,
and the full evaluation protocol are detailed in
Appendix~\ref{app:cityb_oof}.

Table~\ref{tab:cityb_oof} reports the results under this secondary protocol.
L2G-AccountOnly achieves the highest Macro AP (0.6527) and Macro AUROC
(0.6287), whereas L2G-Joint obtains a Macro AP of 0.6372 and a nearly
identical Macro AUROC of 0.6285, together with the highest Localization Macro
AP (0.4985). Operator-specific rankings differ: L2G-Joint achieves the best
K1 AP (0.6847), AnomalyBERT-Proxy performs best on K2 (0.6597), and
NCAD-Native performs best on K3 (0.6763). Thus, no method dominates all
anomaly types, and L2G-Joint combines competitive account-level detection
with the strongest aggregate localization performance.

Several findings from City-A remain observable under this protocol.
Proxy adaptation increases AnomalyBERT's Macro AP from 0.5457 to 0.6263
and NCAD's from 0.6107 to 0.6370, consistent with the benefit of aligning
anomaly supervision with the target deviations. Both L2G variants also
achieve higher Macro AP than Account-Pooled Transformer (0.6078),
supporting the use of visit-level evidence for account-level detection.
However, adding localization supervision increases Localization Macro AP
from 0.3637 to 0.4985 while reducing Macro AP, with almost unchanged
Macro AUROC. The City-B results therefore provide supporting evidence for
anomaly-supervision alignment and local-to-global scoring under the
burden-controlled setting, but do not reproduce the additional account-level
AP benefit of localization supervision observed on City-A.

\section{Conclusion and Future Work}
\label{sec:conclusion}

This paper presents a knowledge-guided local-to-global anomaly learning
framework for longitudinal medication trajectories. By constructing
multi-granularity proxy anomalies with medication knowledge and plausible
negatives, the framework learns anomaly boundaries that distinguish
knowledge-consistent medication variations from suspicious semantic
deviations. The proposed L2G-Joint model further connects visit-level
anomaly evidence with account-level decisions by jointly optimizing
detection and localization through shared visit-level anomaly scores.
Experiments on two independent city cohorts demonstrate the effectiveness of
the proposed framework, with controlled analyses supporting the contributions
of knowledge-guided supervision, plausible-negative construction, and
local-to-global joint learning.

Despite these results, the current framework focuses on detecting and
localizing suspicious deviations based on knowledge-guided proxy anomalies,
which cannot fully determine whether a detected deviation represents a true
clinical anomaly or a complex but reasonable treatment evolution. Future
work will investigate active anomaly verification, where detected deviations
are treated as hypotheses rather than conclusions. A verification agent can
actively gather relevant clinical evidence, examine competing explanations,
and seek evidence that supports or refutes the initial anomaly hypothesis,
moving longitudinal anomaly analysis from detection and localization toward
evidence-based verification.


\appendix

\section{Training and Implementation Details}
\label{app:training_details}

\subsection{Training Configuration}
\label{app:training_config}

\begin{table*}[t]
\centering
\caption{Summary of baseline methods, original formulations, and adaptation
strategies used in our experiments.}
\label{tab:baseline_summary}
\small
\begin{tabular}{lll}
\toprule
Group & Method & Main characteristics \\
\midrule

Native &
Deep SVDD~\citep{ruff2018deepsvdd} &
One-class normality modeling \\

&
NCAD~\citep{carmona2022ncad} &
Contextual anomaly detection with local temporal context \\

&
AnomalyBERT~\citep{jeong2023anomalybert} &
Self-supervised Transformer anomaly detection \\

&
TranAD~\citep{tuli2022tranad} &
Transformer-based reconstruction and anomaly scoring \\

&
CAT~\citep{zhang2022cat} &
Content-aware event sequence anomaly detection \\

&
OC4Seq~\citep{wang2021oc4seq} &
Multi-scale one-class discrete sequence modeling \\

&
EHR-BERT~\citep{niu2024ehrbert} &
Contextual representation learning for EHR sequences \\

&
EHR-HOTN~\citep{niu2022anomalous} &
Higher-order dependency modeling for EHR anomalies \\

\midrule

Proxy-adapted &
Deep SAD~\citep{ruff2020deepsad} &
Semi-supervised anomaly detection \\

&
TAIL-MIL~\citep{jang2025tailmil} &
Time-aware multiple-instance learning \\

&
AnomalyBERT-Proxy~\citep{jeong2023anomalybert} &
AnomalyBERT with knowledge-guided proxy supervision \\

&
NCAD-Proxy~\citep{carmona2022ncad} &
NCAD with knowledge-guided proxy supervision \\

\midrule

Controlled &
Account-Pooled Transformer &
Direct account-level trajectory prediction \\

&
L2G-AccountOnly &
Local evidence learning without localization supervision \\

&
L2G-Joint &
Joint local evidence learning and localization \\

\bottomrule
\end{tabular}
\end{table*}

All trainable models are optimized with a fixed random seed of 42. For the
proposed L2G models, PULSE-derived medication embeddings are kept frozen
during anomaly training and are excluded from optimization. The remaining
model parameters are optimized using AdamW with
$\beta_1=0.9$, $\beta_2=0.999$, and $\epsilon=10^{-8}$.
The batch size is set to 16. The initial learning rates are
$1\times10^{-4}$ for the backbone and $1\times10^{-3}$ for the visit anomaly
head, with a weight decay of $1\times10^{-4}$. Weight decay is disabled for
bias parameters, normalization parameters, and one-dimensional parameters.
Gradient norms are clipped to a maximum global norm of 1.0.

The account-level objective uses an unweighted binary cross-entropy loss.
The final L2G-Joint configuration uses
$\lambda_{\mathrm{local}}=2.0$ and Top-$20\%$ aggregation
($\alpha=0.20$). During training, each training record is used exactly once
per epoch, with deterministic epoch-wise reshuffling.

For the City-A benchmark, checkpoint selection follows a validation-first
protocol based solely on validation Operator-Macro AP. Training initially
uses the original learning rates with an early-stopping patience of five
epochs. The best checkpoint in this stage is obtained at epoch 17. The
complete model state and optimizer state are then restored, and the learning
rates are reduced by a factor of 0.1, resulting in backbone and head learning
rates of $10^{-5}$ and $10^{-4}$, respectively. The final selected checkpoint
is obtained at epoch 18, corresponding to 17 epochs at the initial learning
rates followed by one epoch at the reduced learning rates. A subsequent
low-learning-rate stage with patience of ten does not improve validation
Operator-Macro AP. Validation AUROC and localization AP are recorded only
for analysis and are not used for checkpoint selection. The test set is
accessed only after the final checkpoint is frozen.

For the City-B burden-controlled OOF evaluation, no City-B validation-based
checkpoint selection is performed. Instead, the optimization path selected
from the formal City-A experiment is fixed before evaluation. Each fold is
trained for 17 epochs using the initial learning rates, followed by one epoch
using the reduced learning rates ($0.1\times$). The epoch-18 checkpoint is
then frozen before the corresponding held-out fold is evaluated. No City-B
validation or test result is used to modify training duration or checkpoint
selection.

\subsection{Baseline Adaptation}
\label{app:baseline_adaptation}

We compare the proposed framework with representative anomaly detection
methods covering different anomaly modeling paradigms. Table~\ref{tab:baseline_summary}
summarizes all baseline methods used in our experiments, including their
original formulations and adaptation strategies. The baselines are grouped
into native anomaly methods, proxy-adapted literature methods, and controlled
models according to how anomaly supervision is defined.

Native anomaly methods preserve their original anomaly-learning
formulations and are trained on untouched medication trajectories without
the proposed proxy supervision. This group evaluates whether existing
anomaly definitions developed under their original objectives can transfer
to knowledge-guided medication deviations.

Proxy-adapted literature methods replace their original anomaly sources with
the proposed K1/K2/K3 proxy anomalies while preserving the characteristic
learning mechanisms of each method. This adaptation isolates the effect of
anomaly-supervision alignment from the effect of the proposed local-to-global
architecture. In particular, these methods receive the same suspicious and
plausible proxy trajectories as the proposed framework but retain their
original optimization objectives.

The controlled models are designed to analyze the contribution of local
evidence learning and joint supervision. Account-Pooled Transformer performs
direct account-level prediction from pooled trajectory representations without
explicit visit-level anomaly evidence. L2G-AccountOnly uses the same
visit-level anomaly scoring mechanism and Top-$\alpha$ aggregation as
L2G-Joint but removes localization supervision by setting
$\lambda_{\mathrm{local}}=0$. L2G-Joint further incorporates localization
supervision and jointly optimizes account-level detection and visit-level
localization through shared visit-level anomaly scores.

For methods that naturally produce visit-level anomaly scores but do not
define an account-level anomaly score, the Top-$20\%$ aggregation rule is
used only during evaluation. This evaluation-time aggregation does not modify
the original training objective or model optimization procedure of the
corresponding baseline methods. Methods without visit-level anomaly scores
are reported without localization results.

\subsection{Statistical Analysis}
\label{app:statistical_analysis}

We use paired account-level bootstrap to estimate the uncertainty of
performance differences between compared models. Bootstrap resampling is
performed at the base-account level rather than at the trajectory or
operator-instance level, because multiple proxy trajectories and operator
variants may originate from the same underlying account. Each bootstrap
replicate samples base accounts with replacement while preserving all
corresponding matched proxy instances.

For the main comparisons and aggregation ablations, we use 2,000 bootstrap
replicates with a fixed random seed of 42. Confidence intervals are computed
using the percentile method from the bootstrap distribution. A performance
difference is considered statistically supported when the 95\% confidence
interval does not include zero.

All reported confidence intervals measure uncertainty due to account-level
sampling in the held-out evaluation set. They do not incorporate additional
variation from model initialization or alternative training configurations.
Sensitivity analyses, including $\lambda_{\mathrm{local}}$ and Top-$\alpha$
aggregation studies, are evaluated using the same validation-first protocol
described in the main text.

\section{City-B Burden-Controlled OOF Protocol}
\label{app:cityb_oof}

\subsection{Evaluation Design and Burden Control}
\label{app:cityb_design}

City-B contains a substantially smaller cohort with shorter medication
trajectories than City-A. Under short trajectories, replacing a fixed number
of visits may result in disproportionately large perturbation burdens,
particularly for multi-visit operators such as K2 and K3. To obtain a more
stable cross-city evaluation, we perform a burden-controlled five-fold
out-of-fold (OOF) analysis, where each account serves as held-out data
exactly once. The OOF protocol is designed to improve evaluation stability
under limited cohort size while preserving the original semantic definitions
of the proposed proxy operators.

For an account $i$, let $T_i$ denote the total number of visits. Since the
first visit is used only as historical context, the number of scoreable visits
is defined as $D_i = T_i - 1$.
Let $A_i$ denote the set of scoreable visits directly modified by the proxy
operator. The perturbation burden is defined as $b_i={|A_i|}/{D_i}$.

To prevent excessively strong perturbations caused by short trajectories,
operator-specific burden ceilings are applied. These thresholds are derived
from the 75th percentile of suspicious-positive burden distributions in the
City-A training set and fixed before City-B evaluation: $K1: b_i \leq 0.0769$, 
$K2: b_i \leq 0.3158$,
$K3: b_i \leq 0.4615$.

The burden constraints only exclude overly severe perturbations and do not
modify the underlying semantics of K1, K2, or K3. Specifically, K1 remains a
sparse medication replacement, K2 remains a semantic treatment-episode
replacement, and K3 remains a cross-account segment injection. All original
knowledge constraints, donor restrictions, and affected-visit definitions
are preserved under the burden-controlled protocol.

\subsection{Fold Construction and Training Protocol}
\label{app:cityb_fold}

The 562 eligible City-B accounts are partitioned into five mutually exclusive
outer folds using seed 42. Each account appears exactly once as held-out data,
while the remaining accounts are used for model training and fold-specific
evaluation procedures. All training-dependent preprocessing steps are
performed independently within each fold using only the corresponding
training accounts.

Proxy anomalies are generated after the account-level partition. For each
fold, replacement sources are restricted to the same city and the same data
partition as the recipient account, preventing information leakage across
training, validation, and held-out folds. The original K1, K2, and K3 semantic
constraints, donor selection rules, and affected-visit definitions remain
unchanged except for the burden filtering described above.

Because City-B provides limited operator-specific validation samples, no
City-B validation-based checkpoint selection or hyperparameter tuning is
performed. Training schedules and optimization settings are fixed from the
formal City-A experiments before OOF evaluation. Each fold is trained
independently under this frozen protocol, and held-out evaluation is
performed only after checkpoint freezing.

\subsection{Cross-Fold Evaluation and Metric Aggregation}
\label{app:cityb_metric}

For each held-out fold, evaluation metrics are first computed independently
for each proxy operator. Let $M_{f,o}$ denote the metric value obtained on
fold $f$ for operator $o\in\{K1,K2,K3\}$, and let $n_{f,o}$ denote the number
of matched held-out account pairs available for the corresponding operator.
The operator-level out-of-fold metric is aggregated across folds using
matched-pair counts as weights:

\[
\bar{M}_{o}
=
\frac{
\sum_{f=1}^{5} n_{f,o} M_{f,o}
}{
\sum_{f=1}^{5} n_{f,o}
}.
\]

After obtaining the aggregated metric for each operator, the final
Operator-Macro metric is computed as the unweighted average across K1, K2,
and K3.

This two-stage aggregation ensures that matched-pair counts affect only the
aggregation within the same operator, while different anomaly scales
contribute equally to the final macro metric. The final held-out matched
account counts are 140, 43, and 121 for K1, K2, and K3, respectively.

Raw anomaly scores from different folds are not directly pooled before
computing AP or AUROC. Since each fold is produced by an independently
trained model, score scales may differ across folds; therefore, metrics are
computed within each fold first and aggregated using the above procedure.
The same aggregation strategy is applied to localization metrics, where
available.

\begin{thebibliography}{99}

\bibitem[Bhanja et al.(2026)]{bhanja2026ehrsurvey}
  Soumendra N. Bhanja, Haoran Niu, Yang Chen, Olufemi A. Omitaomu,
  Angela Laurio, Amber Trickey, Vijayalakshmi Sampath, and Jonathan R. Nebeker,
  Emerging anomaly detection techniques for electronic health records: A survey,
  \textit{Intelligence-Based Medicine},
  13:100349,
  2026.

\bibitem[Bouarfa and Dankelman(2012)]{bouarfa2012workflow}
  Loubna Bouarfa and Jenny Dankelman,
  Workflow mining and outlier detection from clinical activity logs,
  \textit{Journal of Biomedical Informatics},
  45(6):1185--1190,
  2012.

\bibitem[Caron et al.(2014)]{caron2014process}
  Filip Caron, Jan Vanthienen, Kris Vanhaecht, Erik Van Limbergen,
  Jochen Deweerdt, and Bart Baesens,
  A process mining-based investigation of adverse events in care processes,
  \textit{Health Information Management Journal},
  43(1):16--25,
  2014.

\bibitem[De Meulemeester et al.(2025)]{demeulemeester2025explainable}
  Hannes De Meulemeester, Frank De Smet, Johan van Dorst,
  Elise Derroitte, and Bart De Moor,
  Explainable unsupervised anomaly detection for healthcare insurance data,
  \textit{BMC Medical Informatics and Decision Making},
  25:14,
  2025.

\bibitem[Edelman et al.(2026)]{edelman2026trajectories}
  Brice Edelman, Hannah Kim, and Jeffrey Skolnick,
  Scalable discovery and validation of order-specific electronic health record
  event trajectories for interpretable adverse-outcome risk estimation,
  \textit{Journal of Biomedical Informatics},
  177:105008,
  2026.

\bibitem[Gebeyehu et al.(2026)]{gebeyehu2026rare}
  Biniyam Gebeyehu, Bennett Kleinberg, Katrijn Van Deun, and Esther de Vries,
  Detection of rare medical events in electronic health records using machine
  learning: Current practices and suggestions---A scoping review,
  \textit{PLOS ONE},
  21(3):e0332963,
  2026.

\bibitem[Hu et al.(2012)]{hu2012utilization}
  Jianying Hu, Fei Wang, Jimeng Sun, Robert Sorrentino, and Shahram Ebadollahi,
  A healthcare utilization analysis framework for hot spotting and contextual
  anomaly detection,
  \textit{AMIA Annual Symposium Proceedings},
  2012:360--369,
  2012.

\bibitem[Huang et al.(2012)]{huang2012clinical}
  Zhengxing Huang, Xudong Lu, and Huilong Duan,
  Anomaly detection in clinical processes,
  \textit{AMIA Annual Symposium Proceedings},
  2012:370--379,
  2012.

\bibitem[Huang et al.(2015)]{huang2015local}
  Zhengxing Huang, Wei Dong, Lei Ji, Liangying Yin, and Huilong Duan,
  On local anomaly detection and analysis for clinical pathways,
  \textit{Artificial Intelligence in Medicine},
  65(3):167--177,
  2015.

\bibitem[Muyama et al.(2024)]{muyama2024clinicalpathways}
  Lillian Muyama, Antoine Neuraz, and Adrien Coulet,
  Machine learning approaches for the discovery of clinical pathways from
  patient data: A systematic review,
  \textit{Journal of Biomedical Informatics},
  160:104746,
  2024.

\bibitem[Niu et al.(2022)]{niu2022anomalous}
  Haoran Niu, Olufemi A. Omitaomu, Michael A. Langston, Mohammad Olama,
  Ozgur Ozmen, Hilda B. Klasky, Angela Laurio, Brian Sauer,
  Merry Ward, and Jonathan Nebeker,
  Detecting anomalous sequences in electronic health records using
  higher-order tensor networks,
  \textit{Journal of Biomedical Informatics},
  135:104219,
  2022.

\bibitem[Niu et al.(2024)]{niu2024ehrbert}
  Haoran Niu, Olufemi A. Omitaomu, Michael A. Langston, Mohammad Olama,
  Ozgur Ozmen, Hilda B. Klasky, Angela Laurio, Merry Ward, and Jonathan Nebeker,
  EHR-BERT: A BERT-based model for effective anomaly detection in electronic
  health records,
  \textit{Journal of Biomedical Informatics},
  150:104605,
  2024.

\bibitem[Perer et al.(2015)]{perer2015care}
  Adam Perer, Fei Wang, and Jianying Hu,
  Mining and exploring care pathways from electronic medical records with
  visual analytics,
  \textit{Journal of Biomedical Informatics},
  56:369--378,
  2015.

\bibitem[Settipalli and Gangadharan(2023)]{settipalli2023wmtdbc}
  Lavanya Settipalli and G. R. Gangadharan,
  WMTDBC: An unsupervised multivariate analysis model for fraud detection in
  health insurance claims,
  \textit{Expert Systems with Applications},
  215:119259,
  2023.

\bibitem[Suesserman et al.(2023)]{suesserman2023procedure}
  Michael Suesserman, Samantha Gorny, Daniel Lasaga, John Helms,
  Dan Olson, Edward Bowen, and Sanmitra Bhattacharya,
  Procedure code overutilization detection from healthcare claims using
  unsupervised deep learning methods,
  \textit{BMC Medical Informatics and Decision Making},
  23:196,
  2023.

\bibitem[Pang et al.(2021)]{pang2021deepad}
  Guansong Pang, Chunhua Shen, Longbing Cao, and Anton van den Hengel,
  Deep learning for anomaly detection: A review,
  \textit{ACM Computing Surveys},
  54(2):1--38,
  2021.

\bibitem[Ruff et al.(2018)]{ruff2018deepsvdd}
  Lukas Ruff, Robert A. Vandermeulen, Nico G{\"o}rnitz, Lucas Deecke,
  Shoaib A. Siddiqui, Alexander Binder, Emmanuel M{\"u}ller, and Marius Kloft,
  Deep one-class classification,
  \textit{Proceedings of the 35th International Conference on Machine Learning},
  80:4393--4402,
  2018.

\bibitem[Shen et al.(2020)]{shen2020thoc}
  Lifeng Shen, Zhuocong Li, and James T. Kwok,
  Timeseries anomaly detection using temporal hierarchical one-class network,
  \textit{Advances in Neural Information Processing Systems},
  33:13016--13026,
  2020.

\bibitem[Su et al.(2019)]{su2019omnianomaly}
  Ya Su, Youjian Zhao, Chenhao Niu, Rong Liu, Wei Sun, and Dan Pei,
  Robust anomaly detection for multivariate time series through stochastic recurrent neural network,
  \textit{Proceedings of the 25th ACM SIGKDD International Conference on Knowledge Discovery \& Data Mining},
  2828--2837,
  2019.

\bibitem[Tuli et al.(2022)]{tuli2022tranad}
  Shreshth Tuli, Giuliano Casale, and Nicholas R. Jennings,
  TranAD: Deep Transformer networks for anomaly detection in multivariate time series data,
  \textit{Proceedings of the VLDB Endowment},
  15(6):1201--1214,
  2022.

\bibitem[Xu et al.(2022)]{xu2022anomalytransformer}
  Jiehui Xu, Haixu Wu, Jianmin Wang, and Mingsheng Long,
  Anomaly Transformer: Time series anomaly detection with association discrepancy,
  \textit{International Conference on Learning Representations},
  2022.
  
  \bibitem[Du et al.(2017)]{du2017deeplog}
  Min Du, Feifei Li, Guineng Zheng, and Vivek Srikumar,
  DeepLog: Anomaly detection and diagnosis from system logs through deep learning,
  \textit{Proceedings of the 2017 ACM SIGSAC Conference on Computer and Communications Security},
  1285--1298,
  2017.

\bibitem[Meng et al.(2019)]{meng2019loganomaly}
  Weibin Meng, Ying Liu, Yichen Zhu, Shenglin Zhang, Dan Pei, Yuqing Liu,
  Yihao Chen, Ruizhi Zhang, Shimin Tao, Pei Sun, and Rong Zhou,
  LogAnomaly: Unsupervised detection of sequential and quantitative anomalies
  in unstructured logs,
  \textit{Proceedings of the Twenty-Eighth International Joint Conference on Artificial Intelligence},
  4739--4745,
  2019.

\bibitem[Wang et al.(2021)]{wang2021oc4seq}
  Zhiwei Wang, Zhengzhang Chen, Jingchao Ni, Hui Liu, Haifeng Chen, and Jiliang Tang,
  Multi-scale one-class recurrent neural networks for discrete event sequence anomaly detection,
  \textit{Proceedings of the 27th ACM SIGKDD Conference on Knowledge Discovery \& Data Mining},
  3726--3734,
  2021.

\bibitem[Zhang et al.(2022)]{zhang2022cat}
  Shengming Zhang, Yanchi Liu, Xuchao Zhang, Wei Cheng, Haifeng Chen, and Hui Xiong,
  CAT: Beyond efficient Transformer for content-aware anomaly detection in event sequences,
  \textit{Proceedings of the 28th ACM SIGKDD Conference on Knowledge Discovery \& Data Mining},
  4541--4550,
  2022.
  
  \bibitem[Guo et al.(2024)]{guo2024logformer}
  Hongcheng Guo, Jian Yang, Jiaheng Liu, Jiaqi Bai, Boyang Wang,
  Zhoujun Li, Tieqiao Zheng, Bo Zhang, Junran Peng, and Qi Tian,
  LogFormer: A pre-train and tuning pipeline for log anomaly detection,
  \textit{Proceedings of the AAAI Conference on Artificial Intelligence},
  38(1):135--143,
  2024.

\bibitem[Guan et al.(2025)]{guan2025feadseq}
  Wei Guan, Jian Cao, Haiyan Zhao, Yang Gu, and Shiyou Qian,
  FeadSeq: A personalized federated anomaly detection framework for discrete
  event sequences,
  \textit{ACM Transactions on Knowledge Discovery from Data},
  19(6):1--26,
  2025.
  
  \bibitem[Hendrycks et al.(2019)]{hendrycks2019oe}
  Dan Hendrycks, Mantas Mazeika, and Thomas G. Dietterich,
  Deep anomaly detection with outlier exposure,
  \textit{International Conference on Learning Representations},
  2019.

\bibitem[Pang et al.(2019)]{pang2019devnet}
  Guansong Pang, Chunhua Shen, and Anton van den Hengel,
  Deep anomaly detection with deviation networks,
  \textit{Proceedings of the 25th ACM SIGKDD International Conference on Knowledge Discovery \& Data Mining},
  353--362,
  2019.

\bibitem[Ruff et al.(2020)]{ruff2020deepsad}
  Lukas Ruff, Robert A. Vandermeulen, Nico G{\"o}rnitz,
  Alexander Binder, Emmanuel M{\"u}ller, Klaus-Robert M{\"u}ller, and Marius Kloft,
  Deep semi-supervised anomaly detection,
  \textit{International Conference on Learning Representations},
  2020.
  
  \bibitem[Goyal et al.(2020)]{goyal2020drocc}
  Sachin Goyal, Aditi Raghunathan, Moksh Jain, Harsha Vardhan Simhadri,
  and Prateek Jain,
  DROCC: Deep robust one-class classification,
  \textit{Proceedings of the 37th International Conference on Machine Learning},
  119:3711--3721,
  2020.

\bibitem[Bergman and Hoshen(2020)]{bergman2020goad}
  Liron Bergman and Yedid Hoshen,
  Classification-based anomaly detection for general data,
  \textit{International Conference on Learning Representations},
  2020.

\bibitem[Li et al.(2021)]{li2021cutpaste}
  Chun-Liang Li, Kihyuk Sohn, Jinsung Yoon, and Tomas Pfister,
  CutPaste: Self-supervised learning for anomaly detection and localization,
  \textit{Proceedings of the IEEE/CVF Conference on Computer Vision and Pattern Recognition},
  9664--9674,
  2021.

\bibitem[Zavrtanik et al.(2021)]{zavrtanik2021draem}
  Vitjan Zavrtanik, Matej Kristan, and Danijel Sko\v{c}aj,
  DRAEM: A discriminatively trained reconstruction embedding for surface anomaly detection,
  \textit{Proceedings of the IEEE/CVF International Conference on Computer Vision},
  8330--8339,
  2021.
  
  \bibitem[Carmona et al.(2022)]{carmona2022ncad}
  Chris U. Carmona, Fran\c{c}ois-Xavier Aubet, Valentin Flunkert, and Jan Gasthaus,
  Neural contextual anomaly detection for time series,
  \textit{Proceedings of the Thirty-First International Joint Conference on Artificial Intelligence},
  2843--2851,
  2022.

\bibitem[Jeong et al.(2023)]{jeong2023anomalybert}
  Yungi Jeong, Eunseok Yang, Jung Hyun Ryu, Imseong Park, and Myungjoo Kang,
  AnomalyBERT: Self-supervised Transformer for time series anomaly detection using data degradation scheme,
  \textit{CoRR},
  abs/2305.04468,
  2023.

\bibitem[Xu et al.(2024)]{xu2024couta}
  Hongzuo Xu, Yijie Wang, Songlei Jian, Qing Liao, Yongjun Wang, and Guansong Pang,
  Calibrated one-class classification for unsupervised time series anomaly detection,
  \textit{IEEE Transactions on Knowledge and Data Engineering},
  36(11):5723--5736,
  2024.

\bibitem[Tang et al.(2024)]{tang2024pcrta}
  Liaoyuan Tang, Zheng Wang, Guanxiong He, Rong Wang, and Feiping Nie,
  Perturbation guiding contrastive representation learning for time series anomaly detection,
  \textit{Proceedings of the Thirty-Third International Joint Conference on Artificial Intelligence},
  4955--4963,
  2024.
  
  \bibitem[Qiu et al.(2024)]{qiu2024augpaste}
  Jiaming Qiu, Weikai Huang, Yijin Huang, Nanxi Yu, and Xiaoying Tang,
  AugPaste: A one-shot approach for diabetic retinopathy detection,
  \textit{Biomedical Signal Processing and Control},
  96:106489,
  2024.

\bibitem[Xia and Mao(2026)]{xia2026eastkb}
  Yushun Xia and Zhizhong Mao,
  Time series anomaly detection based on enhanced anomaly simulation and triplet contrastive knowledge base,
  \textit{Knowledge-Based Systems},
  331:114800,
  2026.
  
  
\bibitem[Jang and Kwon(2025)]{jang2025tailmil}
  Jaeseok Jang and Hyuk-Yoon Kwon,
  TAIL-MIL: Time-Aware and Instance-Learnable Multiple Instance Learning for
  Multivariate Time Series Anomaly Detection,
  \textit{Proceedings of the AAAI Conference on Artificial Intelligence},
  39(17):17582--17589,
  2025.

\end{thebibliography}



\end{document}